\documentclass[11pt]{article}

\usepackage[final]{acl}

\usepackage{fontspec}
\defaultfontfeatures{Ligatures=TeX}
\newfontfamily\devanagarifont[
  Script=Devanagari, Path=fonts/
]{NotoSerifDevanagari-Regular.ttf}

\newfontfamily\tamilfont[
  Script=Tamil, Path=fonts/
]{NotoSerifTamil-Regular.ttf}

\newcommand{\devtext}[1]{{\devanagarifont #1}}
\newcommand{\tamtext}[1]{{\tamilfont #1}}

\newcommand{\emoji}[1]{\ifcsname emojiimage@#1\endcsname\csname emojiimage@#1\endcsname\else\PackageError{gendis}{Unknown emoji #1}{}\fi}
\expandafter\def\csname emojiimage@smiling-face-with-heart-eyes\endcsname{\raisebox{-0.15em}{\includegraphics[height=1em]{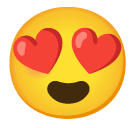}}}
\expandafter\def\csname emojiimage@clapping-hands\endcsname{\raisebox{-0.15em}{\includegraphics[height=1em]{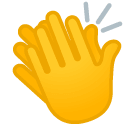}}}
\usepackage{latexsym}
\usepackage{longtable}
\usepackage{array}
\usepackage{booktabs}
\usepackage{multirow}
\usepackage{microtype}
\usepackage{graphicx}
\graphicspath{{images/}}
\usepackage{xcolor}
\usepackage{enumitem}
\usepackage{tabularx}
\usepackage{tikz}
\usepackage{fvextra}
\usepackage{placeins}
\usepackage{dblfloatfix}
\usetikzlibrary{arrows.meta}
\newcolumntype{Y}{>{\raggedright\arraybackslash}X}

\title{\textsc{Gendis}: Gender-Related Discourse and Stance in Multilingual YouTube Comments from India, Kenya, and Nigeria}

\author{
  \textbf{Luise Koch\textsuperscript{1,2}},
  \textbf{Manuel Tonneau\textsuperscript{1,3,4}},
  \textbf{Niyati Malhotra\textsuperscript{1}},
  \textbf{Sunny Rai\textsuperscript{5}},
  \textbf{Ana Maria Munoz Boudet\textsuperscript{1}}, \\
  \textbf{Victor Orozco-Olvera\textsuperscript{1}},
  \textbf{Samuel P. Fraiberger\textsuperscript{1,4}},
  \textbf{Sharath Chandra Guntuku\textsuperscript{5}}
\\
\\
  \textsuperscript{1}World Bank,
  \textsuperscript{2}Technical University of Munich,
  \textsuperscript{3}University of Oxford,
\\
  \textsuperscript{4}New York University,
  \textsuperscript{5}University of Pennsylvania
\\
  \small{
    \textbf{Correspondence:}
    \href{mailto:luise.koch@tum.de}{luise.koch@tum.de}
  }
}

\begin{document}
\maketitle
\begin{abstract}
Research on gender-related online content is often operationalized through harm-centered tasks that collapse non-harmful gender-related and unrelated content, while existing datasets remain concentrated in high-resource languages and Global North settings. To bridge these gaps, we introduce \textsc{Gendis}, a dataset of 9,600 multilingual, human-annotated YouTube comments from India, Kenya, and Nigeria (3,200 per country), annotated using a two-level framework that first identifies whether gender is referenced and then classifies gender-related content as negative, positive, or neutral. This separation provides a denominator for gender-related discourse independent of harm, enabling both prevalence estimation and characterization of its evaluative range. In the uniform random stratum, 23.8--35.0\% of comments are gender-related, and among those with a resolved stance, 82\% are neutral or positive. Across models, gender-reference detection is substantially easier than stance classification, especially for positive and neutral speech. Lightly supervised classifiers match frontier zero-shot large language models at substantially lower cost and outperform generic toxicity detection, while cross-country transfer remains limited. \textsc{Gendis} provides a locally grounded resource for measuring when gender is invoked online and how that reference is framed.
\end{abstract}

\noindent\textbf{Content warning:} This paper contains examples of gender-related online speech that some readers may find offensive or distressing, including sexism, misogyny, transphobia, and references to gender-based violence.

\section{Introduction}

Gender is one of the most common lenses through which people talk about one another online, and the way they invoke it runs a wide range: from misogynistic abuse and policing of how women and men should behave, through plain description, to solidarity, praise, and calls for equality. Each end has real stakes. Hostile gendered attention pushes women and gender minorities out of public conversation \citep{matias2019, koch-etal-2025-tfgbv}, while supportive responses can challenge discriminatory stereotypes, express solidarity, and provide counterspeech \citep{guest-etal-2021-expert,mun-etal-2023-beyond}. 

\begin{figure}[t]
    \centering
    \includegraphics[
        width=\columnwidth,
        trim=2.2cm 15.5cm 3.2cm 2.3cm,
        clip
    ]{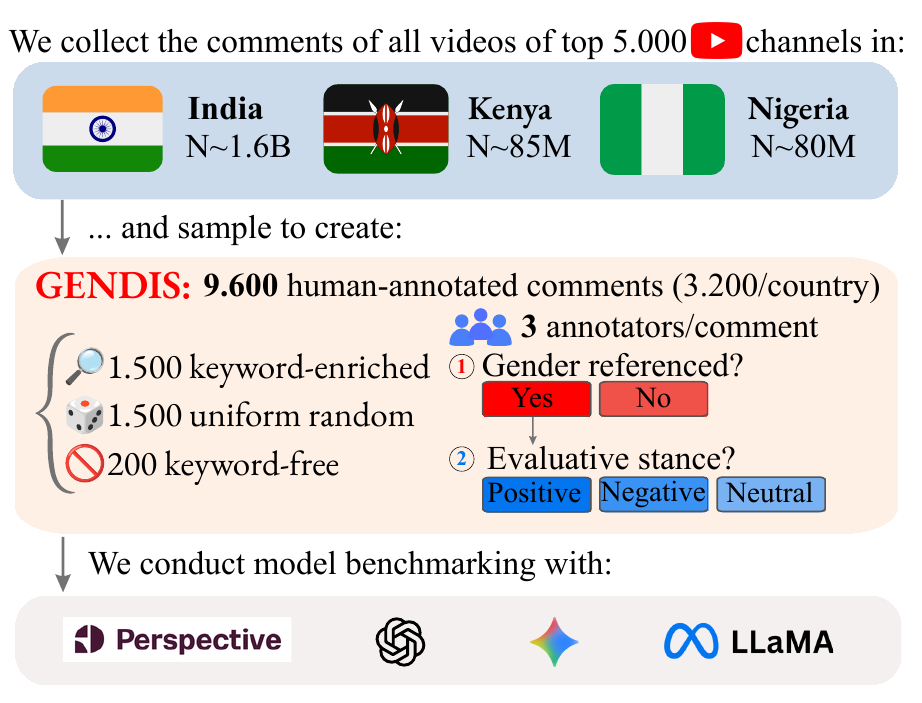}
    \caption{\textsc{Gendis} consists of three annotated datasets
    ($N=3{,}200$ each) sampled from country-associated channel corpora
    in India, Kenya, and Nigeria. The dataset is strongly multilingual
    and code-mixed, with English alongside languages and varieties such
    as Hindi, Swahili, Sheng, Nigerian Pidgin, Yoruba, Hausa, and Igbo.}
    \label{fig:overview}
\end{figure}

Measuring this whole range is therefore necessary not only to identify harmful gender-related speech, but also to characterize what it is being distinguished from: descriptive, supportive, and otherwise non-harmful uses of gender. Similar gendered language can appear in hostile content, neutral description, or counterspeech and solidarity, gender reference, and evaluative stance must be treated as distinct dimensions. Despite this importance, NLP research on gender-related speech faces two limitations. First, gender-related online content has often been operationalized through harm-centered tasks such as sexism and misogyny detection, or within broader hate speech and toxicity frameworks \citep{abercrombie-etal-2023-resources,guest-etal-2021-expert,kirk-etal-2023-semeval}. In such tasks, non-harmful content is often treated as a broad background class, obscuring the distinction between comments unrelated to gender and comments that invoke gender neutrally or positively. Separating gender reference from evaluative stance therefore makes it possible to measure how often gender is invoked independently of harm and, conditional on that reference, how it is framed. Second, existing resources remain concentrated in a limited number of high-resource languages and Global North settings \citep{tonneau-etal-2024-languages}, despite emerging work in settings such as Brazil, India, and Nigeria \citep{arora-etal-2024-uli,tonneau-etal-2024-naijahate,koch-etal-2025-online}. This concentration matters because gendered meanings are expressed through locally specific linguistic varieties, slang, code-switching, and gender norms, and models developed in one national setting often transfer poorly to another \citep{udupa-koch-2024-misogyny, maronikolakis-etal-2022-listening}.

To fill these gaps, we introduce \textsc{Gendis}, a two-level framework and human-annotated evaluation set for gender-related speech in comments from YouTube channels from India, Kenya, and Nigeria (Figure~\ref{fig:overview}). Level~1 of the framework identifies whether a comment is gender-related, and Level~2 classifies its evaluative stance as negative, positive, neutral, or unsure. Using locally reviewed multilingual keyword lists and a random stratum, we sample 9,600 YouTube comments, have them annotated, and evaluate a toxicity baseline, zero-shot large language models, and lightly supervised classifiers on this annotated set at both levels of the framework. We find that detecting gender-related speech is substantially easier than classifying its evaluative stance, where positive and neutral speech are the hardest categories to detect accurately and remain an open challenge. A lightly supervised classifier trained on country-specific annotated data matches frontier zero-shot large language models at both levels at a fraction of the cost and beats the smaller and open models at Level~2, while a generic toxicity classifier performs poorly. We make three contributions:
\begin{enumerate}
\item A two-level annotation framework that first identifies whether gender is referenced and then classifies its evaluative stance, allowing harmful gender-related speech to be distinguished from both gender-unrelated and non-harmful gender-related content.

\item \textsc{Gendis}, comprising locally reviewed multilingual keyword lists and a human-annotated evaluation set of 9,600 comments from country-associated YouTube channel corpora covering India, Kenya, and Nigeria, together with disaggregated annotations and analyses of annotator disagreement.\footnote{Data and code are available at \url{https://github.com/luisekoch/gendis}.}

\item A benchmark of a toxicity baseline, zero-shot large language models, and lightly supervised classifiers, showing that generic toxicity detection is poorly suited to the task and small amounts of country-specific supervision are highly valuable.
\end{enumerate}

\section{Related Work}
\label{sec:related-work}

\begin{table*}[t]
\centering
\scriptsize
\setlength{\tabcolsep}{4pt}
\renewcommand{\arraystretch}{1.05}

\begin{tabularx}{\textwidth}{@{} >{\raggedright\arraybackslash}X >{\raggedright\arraybackslash}p{1.40cm} >{\raggedright\arraybackslash}p{1.90cm} >{\centering\arraybackslash}p{0.60cm} >{\centering\arraybackslash}p{2.55cm} >{\raggedleft\arraybackslash}p{1.00cm} @{}}

\toprule
\textbf{Dataset} &
\textbf{Platform} &
\textbf{Region} &
\textbf{\#Lang} &
\shortstack{\textbf{Gender ref. beyond harm}} &
\textbf{Size} \\
\midrule

EDOS \citep{kirk-etal-2023-semeval}
& Gab/Reddit & Unspecified & 1 & No
& 20{,}000 \\

XtremeSpeech \citep{maronikolakis-etal-2022-listening}
& Multiple & BR/DE/IN/KE & 5 & No
& 20{,}297 \\

NaijaHate \citep{tonneau-etal-2024-naijahate}
& Twitter & NG & 5 & No
& 35{,}976 \\

Uli \citep{arora-etal-2024-uli}
& Twitter & IN & 3 & No
& 23{,}266 \\

AfriHate \citep{muhammad-etal-2025-afrihate}
& Twitter & Africa & 15 & No
& 90{,}437 \\

EXIST 2023 \citep{plaza-etal-2023-exist}
& Twitter & Unspecified & 2 & No
& 10{,}034 \\

\citet{jha-mamidi-2017-compliment}
& Twitter & Unspecified & 1 & No
& 10{,}095 \\

\citet{guest-etal-2021-expert}
& Reddit & Unspecified & 1 & No
& 6{,}567 \\

\citet{singh-etal-2025-misogynistic}
& YouTube & IN & 2 & Partial
& 12{,}698 \\

\citet{bhattacharya-etal-2020-developing}
& YouTube & IN & 3 & Partial
& 12{,}073 \\

ComMA \citep{kumar-etal-2024-comma}
& Multiple & IN & 4 & Partial
& 57{,}363 \\

\midrule

\textbf{\textsc{Gendis} (ours)}
& \textbf{YouTube}
& \textbf{IN/KE/NG}
& \textbf{18}
& \textbf{Yes}
& \textbf{9,600} \\

\bottomrule
\end{tabularx}

\caption{Prior (gendered) harm datasets and \textsc{Gendis} compared across platform, region, number of languages, whether gender is referenced beyond harm, and size. \textsc{Gendis} uniquely treats gender reference beyond harm as a first-class annotation target, capturing negative, positive, and neutral evaluative stance across multiple Global South contexts. Further datasets are discussed in \S\ref{sec:related-work}.}
\label{tab:datasets}
\end{table*}

\paragraph{Annotation of gender-related discourse}
Computational research has largely operationalized gender-related online content through sexism, misogyny, or gender-based violence \citep{abercrombie-etal-2023-resources,guest-etal-2021-expert,kirk-etal-2023-semeval}. Within this harm-centered literature, annotation schemes distinguish increasingly fine-grained forms of hostile and benevolent sexism, pejoratives, misogyny, gendered attacks, and threats \citep{jha-mamidi-2017-compliment, zeinert-etal-2021-annotating, koch-etal-2025-online}. Some schemes further distinguish direct sexism from reported sexism or responses to it \citep{chiril-etal-2020-annotated, plaza-etal-2023-exist}, while others separately annotate gender bias, or optimistic, pessimistic, and neutral attitudes \citep{bhattacharya-etal-2020-developing, kumar-etal-2024-comma, singh-etal-2025-misogynistic}. 
However, these distinctions remain situated within harm-centered schemes or corpora selected around a predetermined gender-related topic. We are not aware of existing resources that combine a first-stage decision about whether a comment invokes gender, regardless of harm, with a subsequent classification of the gender-related content as negative, positive, or neutral.

% Indian resources show the value of code-mixing and locally grounded categories \citep{kumar-etal-2018-aggression,kumar-etal-2024-comma}, and the closest structural precedent annotates Indian English, Hindi, and Bangla YouTube comments at two levels but embeds the gendered label in a task centred on aggression \citep{bhattacharya-etal-2020-developing}. Some schemes separate a sexist utterance from a report of or reaction to it \citep{chiril-etal-2020-annotated,plaza-etal-2023-exist}, but draw this line inside the sexist space rather than first asking whether a comment concerns gender at all. We instead first identify whether a comment is gender-related and then distinguish how gender is used as negative, positive, neutral, or unsure, so prior harm taxonomies inform our negative category without exhausting it. Table~\ref{tab:datasets} contrasts \textsc{Gendis} with these prior resources.

\paragraph{Geographic skew, transfer, and localization} Resources for analyzing socially and culturally situated online discourse remain concentrated in a limited set of languages and national settings \citep{tonneau-etal-2024-languages}. In response to this imbalance, a growing body of locally grounded harmful-language research has incorporated sociocultural knowledge and experience-led annotation in settings including Brazil, India, Kenya, Nigeria, and other African contexts \citep{maronikolakis-etal-2022-listening, koch-etal-2025-online, arora-etal-2024-uli, muhammad-etal-2025-afrihate}. 
This work demonstrates that linguistic forms, annotation judgments, and model performance are sensitive to local cultural and linguistic conditions. However, most of these resources remain focused on hate, abuse, or gender-based harm, and gender-specific resources are usually limited to a single country or linguistic setting. A shared framework for studying broader gender-related discourse across multiple Global South contexts, therefore, remains lacking.

\paragraph{Gendered harm in YouTube comments} YouTube comments are an increasingly common source for studying gendered harm, from misogyny and aggression in Indian languages \citep{bhattacharya-etal-2020-developing} and optimistic, pessimistic, and neutral attitudes in Hindi--English \citep{singh-etal-2025-misogynistic} to two-level sexism schemes for Turkish \citep{mut-altin-saggion-2024-novel} and Italian femicide-related comments, where implicit misogyny is hardest to classify \citep{ferrando-etal-2024-exploring}.\\ 
These resources focus on sexism or misogyny within a single linguistic or national setting. We extend them with a shared two-level framework across India, Kenya, and Nigeria that treats negative, positive, and neutral gender-related speech as substantive categories rather than defining the task around harm. Table~\ref{tab:datasets} contrasts \textsc{Gendis} with these prior resources.

\section{The \textsc{Gendis} Dataset}
\subsection{Large-scale YouTube corpus}
\label{sec:youtube-corpus} 

We used a YouTube corpus constructed from lists obtained through a paid Social Blade subscription\footnote{\url{https://socialblade.com/}}. For each country of interest, namely India, Kenya, and Nigeria, we collected the 5,000 YouTube channels with the highest subscriber counts according to Social Blade at the time of extraction. We subsequently collected publicly available video metadata and comments associated with these channels through YouTube's Data API v3 during the time period ranging from May 2025 to February 2026. The resulting corpus comprises approximately 11 million videos and 1.7 billion comments. It is heavily skewed toward India, which contributes approximately 1.6 billion comments, compared with 85 million for Kenya and 80 million for Nigeria. 

\subsection{Keyword list construction}

\paragraph{Seed lists} To identify comments likely to contain gender-related speech, we constructed country-specific multilingual keyword lists. We reviewed gender-specific datasets and annotation frameworks covering hostile and benevolent sexism \citep{jha-mamidi-2017-compliment}, misogynistic language and counterspeech \citep{guest-etal-2021-expert,zeinert-etal-2021-annotating}, fine-grained sexism \citep{kirk-etal-2023-semeval}, gendered aggression and gender bias \citep{bhattacharya-etal-2020-developing, arora-etal-2024-uli,kumar-etal-2024-comma}, and optimistic, pessimistic, and neutral attitudes toward women \citep{singh-etal-2025-misogynistic}. These resources informed seed categories covering gender-group terms and pronouns, femininity and masculinity, sexuality and bodies, gender roles and behavioral expectations, gendered slurs and abuse, benevolent sexism, equality and supportive language, counterspeech, hashtags, abbreviations, and other gender-relevant expressions. 

\paragraph{Expansion and expert review} The seed lists were expanded in November 2025 using Grok 4 (see full prompt in Appendix~\ref{app:grok-prompt}). One author first back-checked the Grok-generated candidates for duplicates, irrelevant terms, and variants that did not refer to gender. Three country-context reviewers per country then reviewed the lists following written guidelines (Appendix~\ref{app:keyword-review}). Reviewers assessed each term for local relevance and contextual ambiguity, checked translations and language or dialect labels, and identified formal, slang, dialectal, and code-switched usage. They also proposed missing terms and flagged underrepresented forms of gender-related discourse, including vocabulary specific to social groups or online platforms.

\paragraph{Final lists} The final keyword lists comprise 1,530 entries across India (778), Nigeria (386), and Kenya (366). They cover English, relevant local languages and linguistic varieties, code-mixed expressions, and YouTube-specific forms such as hashtags, abbreviations, and emojis. Appendix Table~\ref{tab:keyword-language-coverage} summarizes the principal languages and varieties represented in each list. Of the 1,530 list entries, 776 were observed in the collected YouTube comments (498 in India, 137 in Kenya, and 141 in Nigeria).

\subsection{Sampling}

\label{sec:annotation-sample} For each country, we constructed a sample of 3,200 unique comments in three strata: 1,500 drawn through the country-specific keyword list, 1,500 drawn uniformly at random from the full corpus with no keyword condition, and 200 drawn from comments matching none of the terms in the keyword list.

\paragraph{Keyword stratum} The first stratum was designed to maximize the number of comments likely to contain gender-related speech. For each keyword, we sampled up to 20 comments, retaining all available comments when fewer than 20 were present. We then pooled the keyword-specific samples and deduplicated them by comment ID, so that a comment matching several terms appeared only once. All matched terms were retained in the sampling metadata.

\paragraph{Random stratum} We drew 1,500 comments per country uniformly at random from the full corpus, applying no keyword condition and no other filtering. This keyword-independent stratum avoids keyword-selection bias \citep{wiegand-etal-2019-detection, ousidhoum-etal-2020-comparative} and supports an unbiased estimate of prevalence and of real-world performance, testing whether models recognize gender-related speech rather than reproducing the retrieval terms.

\paragraph{Keyword-free stratum} A third stratum of 200 comments per country was sampled from comments matching no term in the keyword list. It gives annotated coverage of the portion of the corpus the keyword lists do not reach, and tests models on comments that share no lexical overlap with the retrieval terms. Because it describes the keyword-free subpopulation rather than the corpus, we retain it in the dataset and use it in all model evaluation, but exclude it from prevalence estimation, which rests on the uniform random stratum alone.

\subsection{Annotation}
\label{sec:annotation-procedure}

\paragraph{Annotation team} We recruited nine annotators, three per country, through professional networks in India, Kenya, and Nigeria, forming panels with relevant country-specific linguistic, cultural, and social-media knowledge. Annotator demographics, compensation, and linguistic backgrounds are reported in Appendix~\ref{app:annotator-background}.

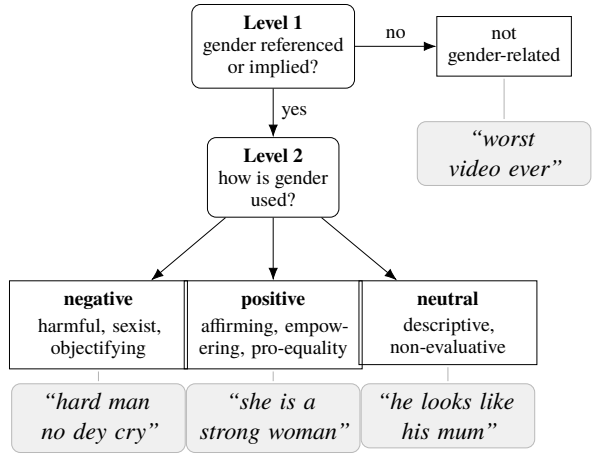
\begin{figure}[t]
\centering
\resizebox{\columnwidth}{!}{%
\begin{tikzpicture}[
  >={Latex[length=2mm]}, font=\footnotesize,
  box/.style={draw, rounded corners, align=center, inner sep=5pt},
  leaf/.style={draw, align=center, inner sep=5pt},
  sub/.style={draw, align=center, text width=2.5cm, anchor=north, inner sep=5pt},
  ex/.style={draw=gray!55, fill=gray!12, rounded corners, align=center,
             text width=2.5cm, anchor=north, inner sep=4pt, font=\itshape},
  link/.style={gray!60, shorten >=1pt, shorten <=1pt}]
  % Level 1
  \node[box]  (l1) at (0,0)      {\textbf{Level~1}\\ gender referenced\\ or implied?};
  \node[leaf] (ng) at (3.7,0)    {not\\ gender-related};
  \node[ex]   (ngex) at (3.7,-1.2) {``worst video ever''};
  \draw[link] (ng) -- (ngex);
  % Level 2
  \node[box]  (l2)  at (0,-2.1)      {\textbf{Level~2}\\ how is gender\\ used?};
  \node[sub]  (neg) at (-2.8,-3.75)  {\textbf{negative}\\[2pt] harmful, \mbox{sexist}, objectifying};
  \node[sub]  (pos) at (0,-3.75)     {\textbf{positive}\\[2pt] affirming, empowering, pro-equality};
  \node[sub]  (neu) at (2.8,-3.75)   {\textbf{neutral}\\[2pt] descriptive, non-evaluative};
  \node[ex]   (negex) at (-2.8,-5.4) {``hard man no dey cry''};
  \node[ex]   (posex) at (0,-5.4)    {``she is a strong woman''};
  \node[ex]   (neuex) at (2.8,-5.4)  {``he looks like his mum''};
  \draw[link] (neg) -- (negex);
  \draw[link] (pos) -- (posex);
  \draw[link] (neu) -- (neuex);
  % flow arrows
  \draw[->] (l1) -- (ng) node[midway, above]{no};
  \draw[->] (l1) -- (l2) node[midway, right]{yes};
  \draw[->] (l2) -- (neg);
  \draw[->] (l2) -- (pos);
  \draw[->] (l2) -- (neu);
\end{tikzpicture}}
\caption{The two-level annotation scheme, with illustrative example comments in the shaded boxes. Level~1 detects whether a comment is gender-related, sending comments with no gender reference (such as general insults) to a leaf and gender-related ones to Level~2, which labels how gender is used. Further examples per category and country are in Appendix~\ref{app:examples}.}
\label{fig:scheme}
\end{figure}

\paragraph{Annotation task} The task follows a two-level structure (Figure~\ref{fig:scheme}). At Level~1, annotators determine whether gender is referenced, explicitly or implicitly. Relevant cues include gendered pronouns and categories, appearance and body references, romantic or family roles, stereotyped femininity or masculinity, gendered slurs, and culturally specific or implied gender meanings. A general insult is not gender-related unless gender is invoked, and if no gender reference is identified, annotation stops at Level~1. At Level~2, for gender-related comments, annotators judge the \emph{evaluative stance}, the way a comment treats gender rather than its surface tone (see Figure~\ref{fig:scheme} and Appendix~\ref{app:examples} for examples). Comments are labeled negative (harmful, stereotyping, objectifying, or violent uses, including benevolent sexism and restrictive gender norms), positive (affirming or empowering a gender group or promoting equality), neutral (descriptive, non-evaluative references, such as identifying ``the woman in blue''), or unsure (gender is present but its stance cannot be determined confidently). Because stance is not tone, benevolent sexism counts as negative despite its positive tone, and where positive and negative content co-occur, the negative label takes precedence. 

\paragraph{Procedure} Each of the 9,600 comments was labeled independently by three annotators in the relevant country panel. Annotators were shown only the text of the sampled comment; video titles, descriptions, surrounding comments, and other contextual metadata were not provided. The guidelines and annotation interface were piloted on a preliminary batch and revised before annotation began, and the task carried a content warning because comments could contain sexism, misogyny, transphobia, sexualized language, and gender-based violence. Annotation was primarily prescriptive: annotators applied the shared definitions, examples, and decision rules of the guidelines (Appendix~\ref{app}) rather than recording purely personal reactions \citep{rottger-etal-2022-two}. Because annotator identities, beliefs, and social positions can nonetheless affect judgments of gender norms and misogyny \citep{sap-etal-2022-annotators, davani-etal-2022-dealing, jiang-etal-2024-examining}, we retain the individual labels and analyze disagreement rather than treating it solely as error.

\paragraph{Label aggregation} At Level~1, the aggregated label is the category selected by at least two of the three annotators. Level~2 ratings are applicable only for annotators who classified the comment as gender-related at Level~1. For a comment with a gender-related Level~1 majority, a resolved Level~2 label requires at least two applicable ratings agreeing on \emph{gender-negative}, \emph{gender-positive}, or \emph{gender-neutral}. Among comments with a gender-related Level~1 majority, 17 had at least two applicable annotators select \emph{unsure} at Level~2 and are reported separately. For the remaining 362 comments, no two applicable annotators agreed on a Level~2 category; we retain these as  \emph{no Level~2 majority} (Appendix Table~\ref{tab:label-distribution}). Neither group serves as a reference class in the benchmark.

\paragraph{Inter-annotator agreement} We calculate inter-annotator agreement from the individual annotations before majority aggregation, over all 3,200 comments per country. Agreement ranges from moderate to substantial and varies by country and annotation level. At Level~1, Krippendorff's $\alpha$ is .80 in India, .79 in Kenya, and .54 in Nigeria. At Level~2, where annotators classify the nature of gendered speech, $\alpha$ is .83 in India, .72 in Kenya, and .49 in Nigeria. For the combined final label, $\alpha$ is .77 in India, .72 in Kenya, and .47 in Nigeria. Fleiss' $\kappa$ gives similar conclusions; full figures are reported in Appendix Table~\ref{tab:iaa}. Level~2 agreement is lower because the task requires annotators not only to identify gender reference, but also to annotate whether the gendered content is negative, positive or neutral. 

\paragraph{Disagreement patterns} Disagreement is concentrated at Level~2 rather than in the basic identification of gender-related speech. Across the annotation, 362 comments received a gender-related Level~1 majority, but no Level~2 majority, meaning that at least two annotators agreed that the comment was gender-related, but no two applicable Level~2 judgements converged on the same negative, positive, or neutral label. This pattern is highly uneven across countries: Nigeria accounts for 306 of these cases, or 18.9\% of its 1,623 gender-related comments, compared with 14 cases in India (1.0\% of 1,470) and 42 in Kenya (3.1\% of 1,350). These splits highlight the interpretive difficulty of Level~2. In many cases, annotators agree that a gendered cue is present but differ over whether the reference is merely descriptive or whether it carries an evaluative, supportive, stereotyping, or derogatory stance. Examples include short, code-mixed, or culturally specific comments built around gendered praise, teasing, or proverbial expressions, such as \emph{nwanyi oma} (Igbo, ``good woman''), ``big boy'', or ``man must hustle''. Such comments can read as a description to some annotators and as an affirmation, stereotype, or normative judgement to others. We therefore treat these cases not simply as annotation noise but as evidence of the normative ambiguity involved in classifying the nature of gendered speech. Following our conservative aggregation rule, comments without a Level~2 majority are assigned to the final unsure category, while a separate provenance flag distinguishes them from comments for which annotators explicitly reached a majority unsure label.

\subsection{Descriptive statistics}
\label{sec:descriptive-statistics}

\paragraph{Label distribution} Across the evaluation set, 4,443 of 9,600 comments (46\%) receive a gender-related Level~1 majority, ranging from 42\% in Kenya to 51\% in Nigeria (Appendix Figure~\ref{fig:distribution}). Among the resolved Level~2 categories, gender-negative is the largest, with 1,868 comments, followed by gender-neutral with 1,621 and gender-positive with 575. Gender-neutral comments are particularly common in Kenya, where they clearly outnumber gender-negative comments (680 versus 503). Nigeria contains both the largest number of gender-negative comments (734) and the largest number of comments with a gender-related Level~1 majority but no Level~2 majority (306). Because the set is keyword-enriched by construction, these shares characterize the annotated sample rather than the prevalence of gender-related speech in the corpus, which we estimate from the random stratum below.

\paragraph{Prevalence} We use the uniform random stratum to estimate the prevalence of gender-related content in comments to top YouTubers per country. About three in ten comments are gender-related, from 23.8\% in Kenya (95\% CI 21.7--26.0) through 30.3\% in India (28.1--32.7) to 35.0\% in Nigeria (32.6--37.4). By comparison, the keyword-enriched stratum is gender-related at 66\%, more than twice as high, confirming that the lists greatly increase collection efficiency. Among the random comments with a resolved Level~2 majority (1,226 of 4,500), 67\% are neutral, 18\% negative, and 16\% positive, so most gender-related speech in this keyword-independent sample is not abusive, the very pattern a harm-only lens misses.

\paragraph{Language composition} To document the linguistic makeup of the annotation sample, we labeled the language of every comment with a large language model (gpt-5-nano, audited in Appendix~\ref{app:language}), recording which languages from a country-specific list appear and whether two or more are mixed in the same comment (Table~\ref{tab:language-composition}). All three sets are strongly multilingual. Code-mixing is pervasive, affecting 32\% of comments in India, 40\% in Kenya, and 30\% in Nigeria. English is widespread, appearing in 58\% of Indian, 67\% of Kenyan, and 82\% of Nigerian comments, alongside major local languages such as Hindi (46\% of Indian comments), Swahili (51\% of Kenyan comments), and Nigerian Pidgin (18\% of Nigerian comments). A further 6 to 12\% of comments are in languages outside each country's list, reflecting the international audiences of large channels. This diversity confirms that English-centric resources would miss much of the gender-related speech in these settings and motivates the locally grounded lists described above.

\paragraph{Gender-positive discourse} To characterize what the positive class contains beyond its aggregate frequency, we conducted an exploratory qualitative analysis of all 575 majority-labeled gender-positive comments. We grouped recurring forms and linguistic patterns descriptively rather than introducing a second annotation scheme; the themes were not independently coded and we therefore do not report inter-annotator agreement or category frequencies. Recurring forms include support for gender equality, safety, and anti-GBV efforts; praise of women’s strength, competence, achievement, and leadership; positive constructions of masculinity emphasizing respect, care, and responsibility; and everyday gendered affirmation. Illustrative country-specific patterns include empowerment and rights discourse in Indian comments, with expressions such as nari shakti (‘women’s power’); Kenyan comments combining expressions such as mwanamke bomba (‘an impressive woman’) and mama wa maendeleo (‘woman of development’) with anti-femicide advocacy and positive constructions of masculinity; and Nigerian comments using vernacular praise forms such as woman wey sabi (‘a capable woman’), nwanyi oma (‘good woman’), and correct guy (‘a good/solid guy’). Positive comments can also refer to harms such as violence, femicide, or abuse while condemning them, showing why positive stance cannot be reduced to positive surface sentiment.

\section{Experiments}

We benchmark a range of models against the human annotations: each reads a comment and predicts whether it is gender-related (Level~1) and, for gender-related comments, its evaluative stance (Level~2). All models are scored on the majority labels over the 9,600 comments from India, Kenya, and Nigeria, letting us compare modeling strategies with very different data and compute requirements.

\subsection{Models}

We benchmark seven models: a training-free toxicity baseline, two supervised classifiers, and four large language models in a zero-shot fashion.

\paragraph{Toxicity baseline} A \emph{toxicity baseline} maps the score of an off-the-shelf multilingual toxicity classifier (\texttt{unitary/multilingual-toxic-xlm-roberta}) onto our labels, treating high-toxicity comments as gender-related and gender-negative, which tests whether gender-related speech reduces to toxicity.

\paragraph{Supervised} We train two supervised classifiers with a single stratified five-fold partition by country and label, reporting out-of-fold predictions so that every comment is scored by a model that never saw it. We fine-tune \textbf{XLM-R} \citep{conneau-etal-2020-unsupervised}, the standard multilingual classification baseline, and train \textbf{SetFit} \citep{tunstall-etal-2022-efficient}, a sample-efficient few-shot method that contrastively fine-tunes a sentence-transformer on same- versus different-label pairs and fits a lightweight classifier on the resulting embeddings, which suits our small data and rare positive and neutral classes. We evaluate three training regimes: within-country (train and test within each country's folds), pooled cross-validation (CV, across all three countries with country-stratified folds), and leave-one-country-out (LOCO, train on two countries and test on the held-out third) to measure cross-cultural transfer.

\paragraph{Zero-shot LLMs} We prompt four large language models, two open and two closed, with the label definitions and a fixed output schema. The \textbf{open} models are the instruction-tuned decoders Llama-3.1-8B and Gemma-2-9B, widely used open-weight reference models at a size that runs on a single GPU and can be deployed locally. The \textbf{closed} models are gpt-5-nano, a small low-cost model of the kind most likely to be deployed at scale, and gpt-5.4, a current frontier model included as a state-of-the-art reference. The prompt was fixed in advance with no prompt search, so the zero-shot scores reflect out-of-the-box behavior.

\subsection{Evaluation}

\paragraph{Metrics} All models are scored on the same instances. Macro-F1 is the headline metric given the class imbalance, reported overall and per country. We evaluate Level~1 (gender-related or not, over all comments), Level~2 (negative, positive, or neutral, conditioned on gold gender-related comments so that Level~1 errors do not leak in), and an end-to-end four-way score (E2E) over the not gender-related, negative, positive, and neutral classes.

\section{Results}

We report the results of our experiments in Table~\ref{tab:results-overall} and Appendix~\ref{app:per-country}. All scores are macro-F1 unless stated otherwise.

\begin{table*}[t]
\centering
\small
\setlength{\tabcolsep}{6pt}
\begin{tabular}{ll c cc cccc}
\toprule
 & & \textbf{Baseline} & \multicolumn{2}{c}{\textbf{Supervised} (CV)} & \multicolumn{4}{c}{\textbf{Zero-shot LLMs}} \\
\cmidrule(lr){3-3}\cmidrule(lr){4-5}\cmidrule(lr){6-9}
 & & Toxicity & XLM-R & SetFit & Llama-3.1-8B & Gemma-2-9B & gpt-5-nano & gpt-5.4 \\
\midrule
\multicolumn{2}{l}{\textbf{Level~1} (detection)} & .46 & .82 & .83 & .76 & .79 & .81 & \textbf{.84} \\
\midrule
\multirow{4}{*}{\textbf{Level~2}}
  & negative & .35 & .71 & .76 & .70 & .71 & .71 & \textbf{.77} \\
  & neutral  & .59 & .55 & \textbf{.66} & .28 & .17 & .45 & .65 \\
  & positive & .00 & .56 & \textbf{.60} & .24 & .34 & .53 & .50 \\
  & macro    & .31 & .61 & \textbf{.67} & .41 & .41 & .56 & .64 \\
\midrule
\multicolumn{2}{l}{\textbf{End-to-end} (4-way)} & .26 & .58 & .63 & .49 & .47 & .61 & \textbf{.64} \\
\bottomrule
\end{tabular}
\caption{F1 on the same $N=9{,}600$ comments for every system, using out-of-fold predictions for the supervised models. Level~1 and End-to-end report macro-F1. The Level~2 rows report per-class F1 and their macro-average, conditioned on gold gender-related comments. Supervised models use pooled cross-validation (CV), with other regimes in Table~\ref{tab:results-regime}. Best per row in bold.}
\label{tab:results-overall}
\end{table*}

\paragraph{Detection easier than stance}
Across every model, macro-F1 is far higher on Level~1 detection than on Level~2 stance classification. SetFit, the strongest model at Level~2, reaches 0.83 on Level~1 but 0.67 on Level~2. Within Level~2, the relative difficulty of the three classes is the same for every model. Every model handles the gender-negative class best and diverges on the other two, with SetFit the only model whose per-class F1 clears 0.55 on all three, and in the end-to-end setting, gender-neutral is the hardest class (per-class F1 0.50 for SetFit CV). This pattern follows the data. Detecting a gender reference can lean on surface cues such as gender-related terms and pronouns, whereas stance must be inferred from context. Neutral, descriptive references sit close to both the not gender-related and gender-negative classes and are most often confused with them (Appendix Figure~\ref{fig:confusion}), while gender-positive comments lack a lexical signature of their own and share the women's-rights and safety vocabulary of the negative pole, so supportive comments are most often misread as negative rather than recognized as affirming.

\paragraph{Supervision rivals frontier models}
The model families separate cleanly, with supervised classifiers matching or beating the zero-shot large language models, which in turn beat the toxicity baseline. SetFit achieves the best macro-F1 at Level~2, within-country or pooled. It ties the zero-shot models on Level~1, and on Level~2 improves over gpt-5-nano by eleven points (0.67 F1 versus 0.56) and over the open decoders by more than twenty-five. Only the frontier gpt-5.4 comes close on this Level~2 gap, reaching 0.64, but it is a large proprietary model billed per call, whereas SetFit trains on country-specific labeled data and runs locally, so a lightly supervised contrastive model matches or beats a frontier zero-shot model at a small fraction of the cost. XLM-R is competitive on Level~1 (0.82) but falls seven points on Level~2, consistent with SetFit being the stronger choice for the rare positive and neutral classes. Among the open decoders, Gemma-2-9B roughly matches gpt-5-nano on Level~1 (0.79 versus 0.81), and Llama-3.1-8B trails only slightly (0.76), so a 9B open model is competitive at detecting gender-related speech, but both fall well behind on Level~2 (0.41 versus 0.56 for gpt-5-nano) and collapse specifically on the neutral class (per-class F1 0.17 and 0.28), plausibly because harm-oriented instruction tuning biases them toward an evaluative reading. The toxicity baseline is weakest of all, lowest on Level~1 (0.46) and never predicting the positive class. Built to fire on abuse, it recovers part of the negative pole but misses neutral and positive gender-related speech, direct evidence that gender-related speech is not reducible to toxicity.

\paragraph{Country and language disparities}
Model performance is uneven across countries and languages, and the two axes reinforce each other. Per-country scores (Appendix Table~\ref{tab:results-country}) do not track annotation agreement: SetFit leads gpt-5-nano on Level~2 in every country, by the widest margin in Nigeria (0.68 versus 0.57), and Nigeria, despite the lowest inter-annotator agreement, has the highest SetFit scores, which we read cautiously because its macro is dominated by the more separable not gender-related and negative comments. Cross-country transfer is the main bottleneck. Within-country training matches pooling all three countries (SetFit within two points on every country), but leave-one-country-out is far worse (0.75 on Level~1, 0.47 on Level~2) and steepest when India is held out (Level~1 0.69). Since a within-country model sees less data than a leave-one-country-out model yet outperforms it, the degradation reflects country-specific distribution shift rather than training-set size (Appendix Table~\ref{tab:results-regime}). The same gap appears across languages (Appendix~\ref{app:results-language}): every model scores highest on English and lower on local-language, code-mixed, and out-of-list comments, widening from Level~1 to Level~2 (for gpt-5-nano, 0.87 on English falling to 0.72--0.76 elsewhere at Level~1, and 0.62 to 0.36 at Level~2). This is not a country artefact, as English outscores non-English within every country. Here too, local supervision helps most, as SetFit is the most robust on out-of-list content (Level~2 0.61 versus 0.36 for gpt-5-nano). Because the language labels are automatic and the smallest group is small, we read these breakdowns as descriptive.

\paragraph{Real-world performance}
Because the headline scores are dominated by the keyword-enriched stratum, we re-score every model on the random stratum alone, drawn uniformly from the corpus regardless of keyword match, which better approximates real-world deployment (Appendix~\ref{app:random-stratum}, Table~\ref{tab:results-random}). Level~1 macro-F1 stays within a point of the full evaluation set for the learned models (SetFit 0.84, gpt-5.4 0.84, gpt-5-nano 0.80), so detecting gender-related speech is not an artefact of the retrieval terms. The toxicity baseline instead collapses to 0.47 and recovers only 7\% of gender-related comments, sharpening the point that gender-related speech is not reducible to toxicity. With round-2 data, Level~2 is reportable on this sample for the first time ($n=1{,}226$, up from 94): SetFit keeps its Level~1 lead and widens it at Level~2 (0.64 versus 0.53 for gpt-5.4), so its stance-classification advantage over the frontier zero-shot model is, if anything, larger on this more representative sample.

\section{Discussion}

\paragraph{Gender-related speech is common and often non-negative}
Sampling independently of our keyword lists, we find that in the uniform random stratum about three in ten comments are gender-related, from 23.8\% in Kenya to 35.0\% in Nigeria, and that most of it is non-negative: among gender-related comments with a resolved stance, 67\% are neutral and 16\% positive. This distribution is not characterized by a harm-centered operationalization \citep{abercrombie-etal-2023-resources,guest-etal-2021-expert,kirk-etal-2023-semeval}, and our toxicity baseline bears it out, recovering part of the negative pole but none of the neutral or positive content that dominates. Still, the negative pole is far from marginal: gender-negative speech runs from 2.5\% of all comments in India to 9.2\% in Nigeria in this keyword-independent sample. This exceeds the hate speech prevalence that \citet{tonneau-etal-2025-hateday} find on a representative day of Twitter, 0.7\% on average and only 0.2\% and 0.3\% in Kenya and Nigeria (1.7\% in India), and most sharply so in Nigeria. Although the platforms differ, the gap suggests that comment sections beneath popular channels concentrate gendered hostility well beyond a platform-wide sample. 

\paragraph{Light local supervision rivals frontier zero-shot models}
A contrastive classifier trained on country-specific labeled data matches the strongest zero-shot large language model at both detecting gender-related speech and judging its stance, and beats the open models by more than twenty points at Level~2, while running locally at a small fraction of the cost. This echoes evidence that lightweight supervised models remain competitive with, or better than, zero-shot large language models for hate speech and sexism detection \citep{tonneau-etal-2025-hateday,plaza-del-arco-etal-2023-respectful}, and the advantage is largest on the rare positive and neutral classes, where general-purpose instruction tuning appears to bias models toward a harm-oriented reading. The qualitative analysis helps explain this difficulty: positive stance ranges from locally specific forms of gendered praise to equality advocacy and anti-GBV solidarity, and the latter often contains the same harm-related vocabulary as negative comments. For a task whose labels are few and locally specific, task-specific supervision is not a fallback but the more accurate and cheaper option wherever per-country data can be collected. Even so, stance classification remains harder than detection for every model, with positive and neutral speech the hardest categories, marking where \textsc{Gendis} leaves the most room for future work.

\paragraph{Gendered meaning is local, and transfer is the bottleneck}
The cost of supervision is that it does not travel. Leave-one-country-out training is far worse than within-country or pooled training even though it uses more data, and every model is strongest on English and weaker on local-language and code-mixed comments, with the gap widening from detection to stance. Because a within-country model outperforms a larger leave-one-country-out one, the degradation reflects country-specific distribution shift rather than data volume. This is consistent with evidence that gendered and harmful language is carried by locally specific varieties, slang, and norms that transfer poorly across settings \citep{tonneau-etal-2024-languages,maronikolakis-etal-2022-listening,koch-etal-2025-online}, and it is why we ground \textsc{Gendis} in three Global South contexts with locally reviewed lists and locally informed annotation rather than porting resources built elsewhere. Measuring gender-related speech across its full range and across settings will require locally grounded data in each new context, not a single model assumed to generalize.

\section{Conclusion}

We introduced \textsc{Gendis}, a two-level framework and human-annotated evaluation set of 9,600 comments from India, Kenya, and Nigeria that separates whether a comment is gender-related from whether the gender-related content is negative, positive, or neutral, treating these as distinct categories rather than collapsing gender into abuse. Agreement varies substantially across the three contexts, and in our benchmark, a contrastive model trained on country-specific labeled data matches frontier zero-shot models at both detecting gender-related speech and classifying it, at a fraction of the cost, while generic toxicity detection fails on both. Much gender-related speech is not abusive, underscoring the need to measure its full evaluative range rather than only its harmful pole. We release \textsc{Gendis} as a foundation for measuring gender-related speech beyond explicit abuse across linguistically and culturally diverse settings.

\section*{Limitations}
The keyword-enriched portion of the evaluation set is built to concentrate gender-related discourse and is not a prevalence estimate: its label distribution describes the annotated sample, not gender-related discourse in the wider corpus. Only the uniform random stratum supports prevalence estimation, and these estimates remain bounded by the Social Blade sampling frame: country assignments do not verify creator nationality, rankings are time-specific, and highly subscribed channels may include organizations, group channels, and international audiences. They therefore characterize the sampled country-associated channel corpora rather than national populations or YouTube users more broadly. The 200 comments per country drawn from keyword-free comments describe that subpopulation rather than the corpus, so we exclude them from prevalence and report them separately. 

Each comment was labeled by three annotators from the relevant country panel. Because these small panels were recruited through professional networks, they may be subject to selection bias and cannot be assumed to represent their national contexts; their size also prevents us from attributing disagreement to particular demographic characteristics. Moreover, the majority labels represent the aggregated judgments of these particular panels under our operational guidelines, not a universal or culturally definitive account of whether gender-related content is positive, negative, or neutral. Other appropriately informed panels could reasonably produce different label distributions, especially for implicit or normatively contested cases.

The study covers three countries and a single platform, so transfer to other regions, languages, or platforms is untested, and our leave-one-country-out results indicate that transfer is already limited across the three countries studied. The country panels did not cover every language and variety present in the sample; although annotators could consult translation tools and flag uncertain cases, linguistic nuance may therefore be lost, particularly in slang and code-mixed content. 

The computational benchmark scores models against aggregated majority labels and excludes comments without a resolved Level~2 label, so it under-represents the most contested cases, which are precisely where measurement is hardest. Finally, we evaluate a single prompt per zero-shot model and one configuration per supervised model, so we do not characterize the prompt and hyperparameter sensitivity that prior work has shown can be substantial.

\section*{Ethics Statement}

The study did not undergo formal institutional ethics review. We nevertheless adopted data-minimization, annotator-welfare, and reporting safeguards appropriate to the potentially harmful and identifiable nature of the material. The study uses comments posted on publicly accessible YouTube pages, but public visibility is not treated as equivalent to informed consent or ethical exemption \citep{franzke-etal-2020-internet}. We reproduce a limited number of short verbatim comments solely to illustrate the annotation categories. Because such comments may remain searchable, we omit usernames, channel and video identifiers, direct links, comment IDs, and other identifying metadata, and avoid reproducing exceptionally sensitive, threatening, or personally identifying material. No such identifiers are included in the illustrative tables.

Annotators were informed that the task could involve sexism, misogyny, transphobia, sexualized language, and references to gender-based violence. They received a content warning, could take breaks, use the unsure category, and contact the research team about difficult or distressing cases. These safeguards recognize that repeated exposure to harmful content can entail emotional labor and psychological burden for annotators and researchers \citep{alemadi-zaghouani-2024-navigating}. Annotators were not required to disclose personal experiences of gender-based harm.

Gender-related speech is culturally and linguistically situated. We recruited annotators with relevant country and language knowledge and collected optional background information to document the perspectives represented in the annotation process. Because the country panels are small, sensitive characteristics are reported only in aggregate and are not used to evaluate individual annotators. Disagreement is retained and analyzed rather than treated solely as
error, but we do not infer that particular demographic characteristics caused particular labels \citep{sap-etal-2022-annotators, davani-etal-2022-dealing}. The authors' own positionality with respect to the settings studied is described in Appendix~\ref{app:positionality}.

The annotated data and keyword lists contain harmful terms and potentially identifying user-generated content. The release of keyword lists will include documentation and a content warning. Raw usernames, direct links, and other identifiers will not be released. The resource is intended for research, measurement, and evaluation. It should not be interpreted as a ready-made system for fully automated moderation, surveillance, or censorship.

Finally, we recognize that research on gendered online abuse can reproduce harm if it extracts from affected communities without accountability. Our analysis therefore focuses on structural patterns, annotation challenges, and measurement limitations rather than sensational examples. The broader aim is to support more accurate and context-sensitive understanding of gendered online harm while acknowledging the limits of computational classification and the responsibilities involved in studying harmful digital content.

\section*{Acknowledgments}
We thank our annotators in India, Kenya, and Nigeria, whose careful work made this dataset possible, and the anonymous reviewers for their constructive feedback.

The study was supported by funding from the Gates Foundation (INV057844), the Penn Global Research Engagement Fund, and Penn MEDIATED. The findings, interpretations, and conclusions expressed in this article are those of the authors alone and shall not be attributed to the Gates Foundation. They also do not necessarily represent the views of the International Bank for Reconstruction and Development/World Bank and its affiliated organizations or those of the Executive Directors of the World Bank or the governments they represent.

\bibliography{custom}

\appendix
\FloatBarrier
\section{Detailed Annotation Guidelines}
\label{app}

This appendix reproduces the guidelines provided to annotators, lightly edited for formatting and to remove identifying details.

\paragraph{Content warning} This document and the related task may include examples of gender-based violence, sexism, transphobia, and other potentially distressing content. Annotators were told to take breaks as needed and to contact the research team if they needed support.

\paragraph{Introduction} Annotators labeled social-media posts (for example, YouTube comments) according to whether they contain gender-related speech and, if so, whether that speech is negative, positive, or neutral.

\paragraph{Labels} Annotators assign at least one primary label per post, in a two-step process. The first step is to identify whether the comment contains gender-related speech. A comment is \emph{gender-related} if it does any of the following:
\begin{itemize}[leftmargin=*]
\item contains the words ``he'', ``his'', ``him'', ``she'', ``her'', or ``hers'' (pay special attention to these words);
\item comments on a person's physical appearance, including wardrobe or pictures;
\item references a person's romantic or sexual relationships, or marital or parental status, for example calling them a mother or father;
\item makes judgments about competence based on a person's gender;
\item references stereotyped masculinity or femininity;
\item mentions a person's gender explicitly, such as stating they are a man, woman, congressman, or congresswoman;
\item encourages the election of a specific gender;
\item compares a person to an individual of the opposite gender;
\item uses gendered slurs or slang such as ``bitch'', ``dick'', ``asshole'', ``dude'', ``yes man'', or ``girl boss'';
\end{itemize}

The second step is to identify the nature of the gender-related comment, as one of four categories:
\begin{itemize}[leftmargin=*]
\item \textbf{Gender positive:} posts that affirm, empower, support, or promote equality for any gender group. This may include gender-equality or rights-based messaging, positive masculinities (emotional openness, non-violence), supportive feminist or pro-equality hashtags (for example \#MeToo, \#HeForShe), and affirmation of trans, nonbinary, or gender-diverse identities.
\item \textbf{Gender negative:} posts that use gender in a harmful, derogatory, stereotyping, exclusionary, sexualizing, objectifying, or violent way. This may include hostile misogyny or misandry, benevolent sexism (paternalistic praise reinforcing hierarchy), restrictive or harmful masculinities, intersectional targeting (gender combined with caste, class, religion, or ethnicity), moral policing or behavioral enforcement, transphobia or gender-policing of LGBTQ+ individuals, and the normalization or trivialization of gender-based violence.
\item \textbf{Neutral:} posts that contain a gender reference that is descriptive, factual, or non-evaluative.
\item \textbf{Flag unsure:} for comments identified as gender-related at Level~1, use this label when the evaluative stance cannot be determined confidently.
\end{itemize}

\paragraph{Decision rules} First, is gender referenced or implied (Level~1)? ``0'' no, not gender-related, stop; ``1'' yes, gender-related, continue. Second, how is gender used (Level~2)? ``0'' descriptive or non-evaluative, neutral; ``1'' harmful or stereotyping, negative; ``2'' affirming or equality-promoting, positive; ``3'' unclear, unsure.

\paragraph{Clarifications}
\begin{itemize}[leftmargin=*]
\item Distinguish clearly between posts that are not gender-related (no reference to gender at all) and those that are gender-related but neutral (gender is mentioned in a purely descriptive, non-evaluative way). If gender is mentioned in any form, treat the post as gender-related and then classify it.
\item Benevolent sexism always counts as gender negative (1).
\item If a post contains both positive and negative gender-related content, classify it as gender negative (1).
\item General insults without gender content are neutral (0).
\item Sexual content counts as gender-related only when it explicitly targets or refers to a specific gender group.
\item Sarcasm reinforcing harmful stereotypes is gender negative (1).
\item Hashtags and emojis count as content and must be labeled accordingly. Ignore URLs.
\item If gender meaning is culturally implied in the local context, treat it as gender-related.
\end{itemize}

\paragraph{Examples}
\begin{itemize}[leftmargin=*]
\item \emph{Gender negative (1):} ``Women shouldn't speak in public.'' ``Real men don't cry.'' ``Trans women are fake.'' ``She should know her place.'' ``Men are trash.''
\item \emph{Gender positive (2):} ``Girls deserve equal opportunities.'' ``Men should be able to express emotions.'' ``Support \#MeToo.'' ``Trans women deserve rights.'' ``Proud to support \#HeForShe.''
\item \emph{Neutral (0):} ``This video is great.'' ``The woman in blue is the host.'' ``Thanks bro'' (non-evaluative use).
\end{itemize}

\paragraph{Edge cases}
\begin{itemize}[leftmargin=*]
\item ``Women are strong, but emotional'' is negative (1), a harmful stereotype.
\item ``Slay queen!'' is positive (2), a positive gender-related expression.
\item ``King behavior!'' is positive (2), positive masculinity.
\item ``Be a real man and stop crying'' is negative (1), restrictive masculinity.
\item ``He's the man of the house'' is negative (1), gender-role enforcement.
\end{itemize}

\paragraph{Annotator tips}
\begin{itemize}[leftmargin=*]
\item If gender is present but its evaluative stance remains unclear, use flag unsure (3) and add a comment.
\item Context matters: rely on cultural interpretation when needed.
\item If a comment contains language you cannot confidently understand, use a translation tool where possible. If the translation remains unclear, particularly for slang, code-mixing, or culturally specific expressions, avoid inferring meaning beyond what can be established from the text; if gender is identifiable but its stance remains unclear, use flag unsure.
\item If gender is explicitly mentioned, the post cannot be neutral unless the reference is purely descriptive.
\item Always prioritize the effect of the statement, not the intention.
\end{itemize}

\onecolumn
\subsection{Additional Illustrative Comments}
\label{app:examples}

Table~\ref{tab:additional-examples} provides two additional
majority-labeled examples of each Level~2 category for each country, complementing the verbatim examples shown in the scheme (Figure~\ref{fig:scheme}). Original spelling, punctuation, grammatical forms, and emojis are preserved. A dash in the translation column indicates that the original comment is already in English. The examples are illustrative and are not intended to represent all gender-related speech within a country.

\begingroup
\footnotesize
\setlength{\tabcolsep}{3pt}
\renewcommand{\arraystretch}{1.12}

\begin{longtable}{@{}
>{\raggedright\arraybackslash}p{1.35cm}
>{\raggedright\arraybackslash}p{1.60cm}
>{\raggedright\arraybackslash}p{6.15cm}
>{\raggedright\arraybackslash}p{6.15cm}
@{}}

\caption{Additional verbatim examples of gender-positive,
gender-neutral, and gender-negative comments from Kenya, Nigeria,
and India.}
\label{tab:additional-examples}\\

\toprule
\textbf{Country} &
\textbf{Label} &
\textbf{Original comment} &
\textbf{English translation} \\
\midrule
\endfirsthead

\multicolumn{4}{l}{
\textit{Table~\ref{tab:additional-examples} continued from the
previous page.}}\\
\toprule
\textbf{Country} &
\textbf{Label} &
\textbf{Original comment} &
\textbf{English translation} \\
\midrule
\endhead

\midrule
\multicolumn{4}{r}{\textit{Continued on the next page.}}\\
\endfoot

\bottomrule
\endlastfoot

% ================================================================
% Kenya
% ================================================================

Kenya &
Positive &
hongera sana dada umeongeza heshima kwa wanawake, kuwa wanaweza &
Congratulations, sister. You have increased respect for women by
showing that women are capable. \\*[\smallskipamount]

&
Positive &
\#endfemicideke...a perfect song for the movement..stop killing women &
-- \\*[\smallskipamount]

&
Neutral &
I didn't know she's the singer of this song , I also like her teachings
pale TikTok &
I didn't know she was the singer of this song. I also like her
teachings on TikTok. \\*[\smallskipamount]

&
Neutral &
cute babies,wanakaa nicholas but rengi ni ya mama yao &
Cute babies, they look like Nicholas, but their complexion comes
from their mother. \\*[\smallskipamount]

&
Negative &
kuna baadh ya wanawake wana makalio makubwa ila brain hakuna kitu &
Some women have big buttocks, but no brains. \\*[\smallskipamount]

&
Negative &
wanawake wenye sauti kama hiyo are very rudy &
Women with voices like that are very rude. \\

\midrule

% ================================================================
% Nigeria
% ================================================================

Nigeria &
Positive &
we should all be feminists - see the ted talk by chimamanda ngozi
adichie. &
-- \\*[\smallskipamount]

&
Positive &
You're doing a great job Abigail! I'm a fellow Nigerian go girl! &
-- \\*[\smallskipamount]

&
Neutral &
wai mumuni matan shi na waneh da allah &
For God's sake how many wives does the holyman have? \\*[\smallskipamount]

&
Neutral &
from a nigerian to you; 'aristo' is a slang for sugar daddy - a rich
man, usually more mature in age who is probably married and has a side
chick(a young woman/girl) and spends money on them. &
-- \\*[\smallskipamount]

&
Negative &
nice movie. i almost cried but hard man no dey cry. i give it
10/10\emoji{smiling-face-with-heart-eyes}%
\emoji{smiling-face-with-heart-eyes} &
Nice movie. I almost cried, but a tough man does not cry.
I give it 10/10.\emoji{smiling-face-with-heart-eyes}%
\emoji{smiling-face-with-heart-eyes} \\*[\smallskipamount]

&
Negative &
this girl is insane. need to repent her behavior is bad. 21 years with
fallen breasts and moving naked in the house. a woman must respect her
body period &
-- \\

\midrule

% ================================================================
% India
% ================================================================

India &
Positive &
mouli doesnt have to pretend... she is a strong women &
-- \\*[\smallskipamount]

&
Positive &
\tamtext{பெண்களுக்கும் அரசியல் அதிகாரத்தை சரிபாதியாக கொடுப்பது
பெண் உரிமை . நாம் தமிழர் கட்சி சீமான்} &
Giving women an equal share of political power is a women's right.
Naam Tamilar Katchi, Seeman. \\*[\smallskipamount]

&
Neutral &
Harsh bhaiya ko bulawo &
Call Harsh bhaiya [older brother]. \\*[\smallskipamount]

&
Neutral &
teri bhabhi &
Your sister-in-law. \\*[\smallskipamount]

&
Negative &
\devtext{इससे यह पता चलता है कि लड़कियों के दिमाग घुटने में ही होते हैं।} &
This shows that girls' brains are in their knees, meaning that girls
are unintelligent. \\*[\smallskipamount]

&
Negative &
ise feminism ki bimari h aur kuch nhi. famous bhi hona h to mens ki
burai karke. wah wah \emoji{clapping-hands}\emoji{clapping-hands} &
She has the disease of feminism, nothing else. She also wants to
become famous by speaking badly about men. Wow. wah \emoji{clapping-hands}\emoji{clapping-hands} \\

\end{longtable}

\endgroup
\twocolumn

\FloatBarrier
\section{Keyword Resource Review and Construction}
\label{app:keyword-construction}

Country-context reviewers reviewed the country-specific lists for local relevance, translation accuracy, language and dialect labels, and formal, slang, hybrid, or code-switched usage. They were also asked to identify ambiguous or rarely used terms and propose missing expressions, particularly coded language and locally specific slang.

\subsection{Keyword expansion prompt}
\label{app:grok-prompt}

The following shared prompt template was used to expand the seed
lists. The country, relevant languages and linguistic varieties,
and attached seed list were changed for each context. Follow-up prompts
were issued separately for individual languages and varieties.

\begin{Verbatim}[
  breaklines=true,
  breakanywhere=true,
  fontsize=\small
]
I am reaching out to ask for your support in reviewing and refining
the attached list of keywords related to gender-based violence and
gendered hate speech.

Focus on keywords used in [COUNTRY]. Generate terms both in English
transliteration and in the relevant native script. Use information
from multiple social-media platforms and the wider internet.

Assess the relevance, accuracy, and cultural appropriateness of the
terms, with particular attention to how gender-based violence and
gendered language appear in YouTube content.

Please review the attached instructions and guideline document.
Do not impose a strict limitation on the range of search terms or
categories. Be expansive and provide a list of 100 candidate terms
across different categories.
\end{Verbatim}

For Kenya, the following country-specific follow-up instruction was
used:

\begin{Verbatim}[
  breaklines=true,
  breakanywhere=true,
  fontsize=\small
]
Kenyan keywords are attached in a text file. Expand the list to cover
Swahili, Sheng, English, and other relevant Kenyan languages or
dialects. Be as exhaustive as possible and provide a full list.
\end{Verbatim}

Equivalent follow-up prompts were issued for India and Nigeria, with the relevant country, language, and linguistic-variety names
substituted. Additional prompts requested the expansion separately for each language or variety.

\subsection{Keyword-list review guidelines}
\label{app:keyword-review}

Reviewers completed three tasks. First, they assessed whether each candidate term was actively used in their local context, flagged terms that were rare or ambiguous without context, and identified missing terms. Second, they checked translations and language or dialect labels, including whether terms were formal, slang, hybrid, or code-switched. Third, they proposed additional locally used keywords, supplying a translation or explanation and a language or dialect label where possible. Reviewers were also asked whether particular forms of gender-related speech (including positive gender-related speech, benevolent sexism, transphobia, and economic control) or vocabulary associated with particular social groups and platforms were underrepresented.

\begin{table*}[t]
  \centering
  \small
  \setlength{\tabcolsep}{5pt}
  \renewcommand{\arraystretch}{1.12}
  \begin{tabularx}{\textwidth}{@{}l Y r@{}}
    \toprule
    \textbf{Country} & \textbf{Principal languages and varieties represented} & \textbf{Entries} \\
    \midrule
    India & English, Hindi, Bengali, Marathi, Tamil, Telugu, Hinglish, Sylheti, Urdu-linked expressions, and other code-mixed forms & 778 \\
    Nigeria & English, Nigerian Pidgin, Hausa, Igbo, Yoruba, and code-mixed forms & 386 \\
    Kenya & English, Swahili, Sheng, Luo, Kikuyu, Meru, Embu, Kalenjin, and code-mixed forms & 366 \\
    \midrule
    \textbf{Total} & & \textbf{1,530} \\
    \bottomrule
  \end{tabularx}
  \caption{Principal language and variety coverage of the final country-specific keyword lists. The lists also include hashtags, abbreviations, emojis, transliterations, slang, and other platform-specific or hybrid forms.}
  \label{tab:keyword-language-coverage}
\end{table*}

\FloatBarrier
\section{Language Composition}
\label{app:language}

To characterize the linguistic makeup of the annotation sample, we labeled the language of each of the 9,600 comments with a large language model (gpt-5-nano), prompted separately for each country with a list of the locally relevant languages plus an ``Other'' option for any language outside that list. For each comment, the model returned the set of languages present and a flag indicating whether two or more of them are mixed. We take the code-mixing rate from this flag and report each language as a presence rate, the share of comments in which it appears, including within code-mixed comments. Because a single comment can contribute to more than one language, presence rates sum to more than 100\%. Table~\ref{tab:language-composition} summarizes the result.

These labels are automatic and approximate. Spot-checking indicates reliable identification of English, the dominant local languages, and out-of-list languages such as Russian and Arabic, but a fuzzier boundary between Swahili, Sheng, and Somali in Kenya, and occasional romanized local-language comments misrouted to ``Other''.
% TODO (author reminder, not for the compiled PDF): have a research assistant audit these automatic language labels on the 50-comment per-country validation samples (shuffled and stratified by predicted category), and update Table~\ref{tab:language-composition} if the audit reveals systematic error, especially at the Sheng/Swahili boundary and for romanized local languages.

\begin{table}[t]
\centering
\small
\begin{tabular}{lr}
\toprule
\textbf{Language / category} & \textbf{\% of comments} \\
\midrule
\multicolumn{2}{l}{\textit{India}} \\
\quad English & 58 \\
\quad Hindi & 46 \\
\quad Tamil & 11 \\
\quad Bengali & 5 \\
\quad Telugu & 3 \\
\quad Marathi & 2 \\
\quad Code-mixed & 32 \\
\quad Other (off-list) & 6 \\
\addlinespace
\multicolumn{2}{l}{\textit{Kenya}} \\
\quad English & 67 \\
\quad Swahili & 51 \\
\quad Sheng & 9 \\
\quad Somali & 5 \\
\quad Code-mixed & 40 \\
\quad Other (off-list) & 12 \\
\addlinespace
\multicolumn{2}{l}{\textit{Nigeria}} \\
\quad English & 82 \\
\quad Nigerian Pidgin & 18 \\
\quad Yoruba & 12 \\
\quad Hausa & 7 \\
\quad Igbo & 7 \\
\quad Code-mixed & 30 \\
\quad Other (off-list) & 10 \\
\bottomrule
\end{tabular}
\caption{Language composition of the annotation sample, as a share of each country's 1,000 comments. Values are presence rates (a comment can contribute to more than one language, so they sum to over 100\%). \emph{Code-mixed} is the share of comments mixing two or more languages, and \emph{Other} the share in languages outside the country's language list. Languages present in under 1.5\% of comments are omitted. Labels are automatic and approximate (Appendix~\ref{app:language}).}
\label{tab:language-composition}
\end{table}

\FloatBarrier
\section{Annotator Background}
\label{app:annotator-background}

\paragraph{Demographics} The nine annotators (three per country) were each a national of the country whose comments they annotated. Across the three teams, six were women and three were men, with one man and two women in each country. The Indian team was aged 20 to 29: two members had completed undergraduate degrees and were pursuing postgraduate study, and one was an undergraduate student. The Kenyan team had one member aged 20 to 29 and two aged 30 to 39, all educated to at least undergraduate level, with degrees in fields such as engineering and law. The Nigerian team had two members aged 30 to 39 and one aged 40 to 49, with backgrounds spanning computer science, business and marketing, and medicine and global health.

\paragraph{Compensation} Annotators were compensated according to institutional short-term consultant rates set by country and consultant level. Because compensation is a central dimension of ethical annotation work, and dataset documentation should make the annotation labor and pay conditions transparent, we report these rates rather than only stating that annotators were paid \citep{silberman-etal-2018-responsible, shmueli-etal-2021-beyond}. Daily rates were INR 400 in India, KES 1,360 to 1,808 in Kenya, and NGN 11,775 in Nigeria, corresponding to approximately USD 4.40, USD 10.47 to 13.92, and USD 8.60 per day, respectively. The annotation task took approximately 10-12 hours per annotator.

\paragraph{Reporting} Background information was provided voluntarily and was available for eight of the nine annotators, with the ninth annotator's characteristics reported from the research team's knowledge. We report characteristics at the level of each country team rather than for individual annotators, because the small panels could otherwise make annotators identifiable. Although annotator backgrounds may shape the interpretation of culturally specific and normatively ambiguous comments, the small panels do not allow us to attribute disagreement to particular demographic characteristics.

\begin{table*}[t] \centering \small \setlength{\tabcolsep}{5pt} \renewcommand{\arraystretch}{1.15} \begin{tabularx}{\textwidth}{@{}p{1.6cm}p{1.6cm}p{4.8cm}Y@{}} \toprule \textbf{Context} & \textbf{Responses} & \textbf{Reported languages and varieties} & \textbf{Relevant contextual familiarity} \\ \midrule India & 3/3 & English, Hindi, and Bengali, with additional partial familiarity with other Indian languages reported. & Respondents reported familiarity with Indian social-media discourse, gendered cultural references, and relevant stereotypes. \\ Nigeria & 3/3 & English, Hausa, and Yoruba. & Respondents reported familiarity with Nigerian social-media discourse and locally used gendered expressions. \\ Kenya & 2/3 & English, Swahili/Kiswahili, Sheng, Kikuyu, Kamba, and Luo. & Respondents reported familiarity with Kenyan social-media discourse, code-switching, and locally used gendered expressions. \\ \bottomrule \end{tabularx} \caption{Aggregated linguistic and contextual expertise reported by the eight annotators who completed the optional questionnaire. Sensitive demographic and professional characteristics are omitted because the country-level panels are small.} \label{tab:annotator-background} \end{table*}

\FloatBarrier
\section{Author Positionality}
\label{app:positionality}
The author team spans academic institutions in Europe and the United States and an international development organization, with disciplinary backgrounds in gender and technology-facilitated violence, computational social science, NLP, and development research. Several authors have personal, academic, or professional ties to India, and authors familiar with the Indian context contributed to reviewing and refining the corresponding keyword resources. We do not assume that the research team itself represents the cultural and linguistic diversity of the three country contexts studied. We therefore relied on country-context reviewers and annotators in India, Kenya, and Nigeria for local linguistic and cultural interpretation, while recognizing that these small panels likewise cannot represent their national contexts in full.

\FloatBarrier
\section{Label Distribution and Inter-Annotator Agreement}
\label{app:labels-agreement}

\begin{figure}[t]
\centering
\includegraphics[width=\columnwidth]{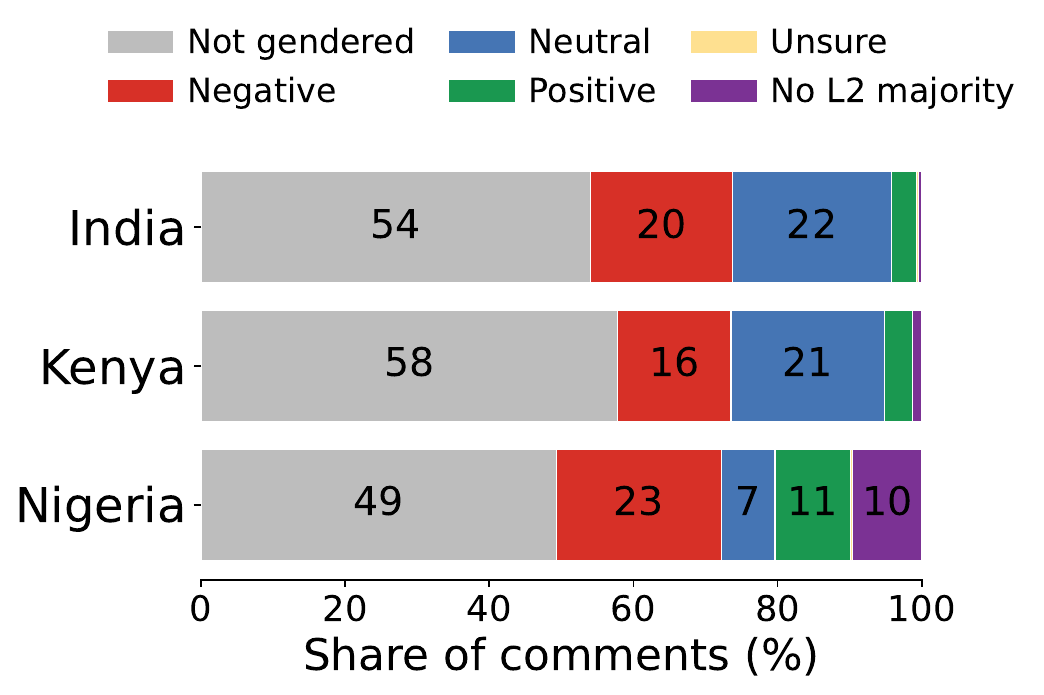}
\caption{Distribution of aggregated labels within each country's 3,200-comment annotated sample. Each sample comprises 1,500 keyword-enriched comments, 1,500 comments drawn uniformly at random from the full corpus, and 200 keyword-free comments. Because the full sample is partly keyword-enriched, these percentages do not estimate prevalence in the broader corpus; prevalence estimates use the uniform random stratum only. Exact counts are reported in Table~\ref{tab:label-distribution}.}
\label{fig:distribution}
\end{figure}

Table~\ref{tab:label-distribution} reports the distribution of majority Level~1 and Level~2 labels for each country, and Figure~\ref{fig:distribution} visualizes the aggregated label shares within each 3,200-comment sample. Table~\ref{tab:iaa} reports inter-annotator agreement as Fleiss' $\kappa$ and Krippendorff's $\alpha$ at both levels and for the combined label. Agreement varies across the three countries and is consistently lower at Level~2 than Level~1, reflecting that judging whether the gender-related content is negative, positive, or neutral is harder than judging whether gender is present.

\begin{table*}[t]
  \centering
  \small
  \begin{tabular}{lrrrrrr}
    \toprule
    \textbf{Country} & \textbf{Not gender-related} & \textbf{GENNEG} & \textbf{GENPOS} & \textbf{Neutral} & \textbf{Unsure} & \textbf{No L2 majority} \\
    \midrule
    India   & 1{,}730 & 631 & 113 & 704 & 8 & 14 \\
    Kenya   & 1{,}850 & 503 & 125 & 680 & 0 & 42 \\
    Nigeria & 1{,}577 & 734 & 337 & 237 & 9 & 306 \\
    \midrule
    All     & 5{,}157 & 1{,}868 & 575 & 1{,}621 & 17 & 362 \\
    \bottomrule
  \end{tabular}
  \caption{Distribution of majority labels by country ($N = 3{,}200$ each). GENNEG denotes gender-negative and GENPOS gender-positive comments. A Level~2 label is assigned when at least two of the three annotators agree on it. \emph{Unsure} counts comments whose agreed Level~2 label is unsure, and \emph{No L2 majority} counts gender-related comments on which no two annotators agree on any Level~2 label. The two columns are reported separately and under the same rule for all three countries.}
  \label{tab:label-distribution}
\end{table*}

\begin{table*}[t]
  \centering
  \small
  \setlength{\tabcolsep}{5pt}
  \begin{tabular}{lccc}
    \toprule
    \textbf{Country} &
    \textbf{Level~1} &
    \textbf{Level~2} &
    \textbf{Combined} \\
    \midrule
    India   & .80 / .80 & .82 / .83 & .77 / .77 \\
    Kenya   & .79 / .79 & .71 / .72 & .72 / .72 \\
    Nigeria & .55 / .54 & .49 / .49 & .47 / .47 \\
    \midrule
    All     & .71 / .71 & .70 / .68 & .66 / .65 \\
    \bottomrule
  \end{tabular}
  \caption{Inter-annotator agreement by country, reported as Fleiss' $\kappa$ /
  Krippendorff's $\alpha$ and calculated from the individual annotations before
  majority aggregation, over all 3,200 comments per country and all 9,600
  comments for the All row. Level~1
  distinguishes gendered from not-gendered comments. Level~2 distinguishes
  negative, positive, neutral, and unsure gendered speech. The combined label
  treats not-gendered comments as one category and combines gendered comments
  with their Level~2 label.}
  \label{tab:iaa}
\end{table*}

\FloatBarrier
\section{Per-Country Model Results}
\label{app:per-country}

Table~\ref{tab:results-country} reports SetFit macro-F1 per country under the three training regimes, with zero-shot gpt-5-nano for comparison, and Table~\ref{tab:results-regime} reports overall macro-F1 for both supervised models under each regime. They back the cross-cultural-transfer analysis in the main text: within-country and pooled CV are close and both far exceed leave-one-country-out, so the transfer drop reflects cross-cultural difference rather than training-set size.

\begin{table}[t]
\centering
\small
\setlength{\tabcolsep}{4pt}
\begin{tabular}{lcccc}
\toprule
 & \multicolumn{3}{c}{\textbf{SetFit}} & \\
\cmidrule(lr){2-4}
\textbf{Country} & \textbf{Within} & \textbf{CV} & \textbf{LOCO} & \textbf{gpt-5-nano} \\
 & {\footnotesize L1 / L2} & {\footnotesize L1 / L2} & {\footnotesize L1 / L2} & {\footnotesize L1 / L2} \\
\midrule
Kenya   & .86 / .64 & .84 / .64 & .75 / .48 & .84 / .56 \\
Nigeria & .86 / .69 & .86 / .68 & .81 / .41 & .80 / .57 \\
India   & .79 / .64 & .79 / .64 & .69 / .48 & .78 / .55 \\
\bottomrule
\end{tabular}
\caption{Per-country macro-F1 for SetFit under the three training regimes, with zero-shot gpt-5-nano for comparison. Within-country and pooled CV are within two points of each other in every country and both far exceed leave-one-country-out, whose drop is steepest for India, so the degradation reflects cross-cultural difference rather than training-set size.}
\label{tab:results-country}
\end{table}

\begin{table}[t]
\centering
\small
\begin{tabular}{llccc}
\toprule
\textbf{System} & \textbf{Regime} & \textbf{L1} & \textbf{L2} & \textbf{E2E} \\
\midrule
SetFit & Within & \textbf{.84} & \textbf{.70} & \textbf{.65} \\
       & CV     & .83 & .67 & .63 \\
       & LOCO   & .75 & .47 & .46 \\
\midrule
XLM-R  & Within & \textbf{.82} & .58 & .55 \\
       & CV     & \textbf{.82} & \textbf{.61} & \textbf{.58} \\
       & LOCO   & .77 & .49 & .48 \\
\bottomrule
\end{tabular}
\caption{Supervised models by training regime: within-country, pooled cross-validation (CV), and leave-one-country-out (LOCO), reported as overall macro-F1. Best regime per model and metric in bold: within-country and CV trade places and are always close, while LOCO is never best, the signature of a transfer gap. Per-country SetFit results are in Table~\ref{tab:results-country}.}
\label{tab:results-regime}
\end{table}

\FloatBarrier
\section{Detection on the Random Stratum}
\label{app:random-stratum}

The main evaluation set is dominated by the keyword-enriched stratum, which by construction concentrates gender-related speech. To gauge detection on a sample closer to unrestricted comments, we re-score every model on the random stratum alone (comments drawn uniformly from the corpus regardless of keyword match, $n=4{,}500$, of which 1,337 are gender-related and 1,226 carry a resolved Level-2 label). Table~\ref{tab:results-random} gives the results. Every learned model stays within a point of its full-set Level~1 score, so detection is not inflated by the retrieval terms, whereas the toxicity baseline falls to 0.47 and recovers only 7\% of gender-related comments. Level~2 is reportable on this sample for the first time: SetFit keeps its Level~1 lead and extends it at Level~2 (0.64 versus 0.53 for gpt-5.4).

\begin{table}[!t]
\centering
\small
\begin{tabular}{lcc}
\toprule
\textbf{System} & \textbf{L1} & \textbf{L2} \\
\midrule
SetFit (CV) & \textbf{.84} & \textbf{.64} \\
gpt-5.4 & \textbf{.84} & .53 \\
XLM-R (CV) & .83 & .58 \\
gpt-5-nano & .80 & .51 \\
Gemma-2-9B & .76 & .31 \\
Llama-3.1-8B & .73 & .35 \\
Toxicity & .47 & .38 \\
\bottomrule
\end{tabular}
\caption{Macro-F1 on the random stratum only ($n=4{,}500$, of which 1,226 gender-related with a resolved Level-2 label), a keyword-independent sample. Level~1 scores stay within a few points of the full evaluation set (Table~\ref{tab:results-overall}), so keyword enrichment does not overstate detection, while the toxicity baseline collapses. With round-2 data, Level~2 is reportable here for the first time (previously $n=94$): SetFit keeps its lead, while gpt-5.4's Level~2 score drops well below its keyword-enriched score, widening the gap with SetFit on this more representative sample.}
\label{tab:results-random}
\end{table}

\FloatBarrier
\section{Performance by Language}
\label{app:results-language}

Using the automatic language labels (Appendix~\ref{app:language}), we group the comments into English, local-language (non-English monolingual), code-mixed, and out-of-list (Other) sets, and recompute macro-F1 within each group (Table~\ref{tab:results-language}). Every model is strongest on English and weaker elsewhere, the gap widens from Level~1 to Level~2, and the locally supervised SetFit is the most robust on non-English and code-mixed content. As these labels are automatic and approximate (Appendix~\ref{app:language}), and the Other group is small, the breakdown is descriptive.

\begin{table*}[t]
\centering
\small
\setlength{\tabcolsep}{6pt}
\begin{tabular}{lcccccccc}
\toprule
& \multicolumn{4}{c}{\textbf{Level~1 (detection)}} & \multicolumn{4}{c}{\textbf{Level~2 (stance)}} \\
\cmidrule(lr){2-5}\cmidrule(lr){6-9}
\textbf{System} & Eng & Local & Mixed & Other & Eng & Local & Mixed & Other \\
\midrule
SetFit (CV) & .90 & .73 & .79 & .71 & .71 & .62 & .65 & .61 \\
gpt-5-nano & .87 & .76 & .75 & .72 & .62 & .51 & .55 & .36 \\
Llama-3.1-8B & .83 & .69 & .71 & .70 & .47 & .34 & .40 & .26 \\
Gemma-2-9B & .85 & .73 & .73 & .73 & .44 & .40 & .38 & .29 \\
XLM-R (CV) & .89 & .71 & .77 & .68 & .62 & .56 & .58 & .58 \\
Toxicity & .50 & .40 & .41 & .51 & .36 & .24 & .32 & .35 \\
\bottomrule
\end{tabular}
\caption{Macro-F1 by language group for Level~1 (gender-related or not) and Level~2 (negative, positive, or neutral, conditioned on gold gender-related comments). Groups are derived from the automatic per-comment language label: \emph{Eng}lish, \emph{Local} (non-English monolingual), \emph{Mixed} (code-mixed), and \emph{Other} (languages outside the country's language list). Group sizes: English 3{,}509, Local 2{,}078, Code-mixed 3{,}287, Other 726. Every system peaks on English and the gap widens at Level~2, while SetFit holds up best on the Other group.}
\label{tab:results-language}
\end{table*}

\FloatBarrier
\section{Model Error Analysis}
\label{app:confusion}

\begin{figure}[t]
\centering
\includegraphics[width=0.85\columnwidth]{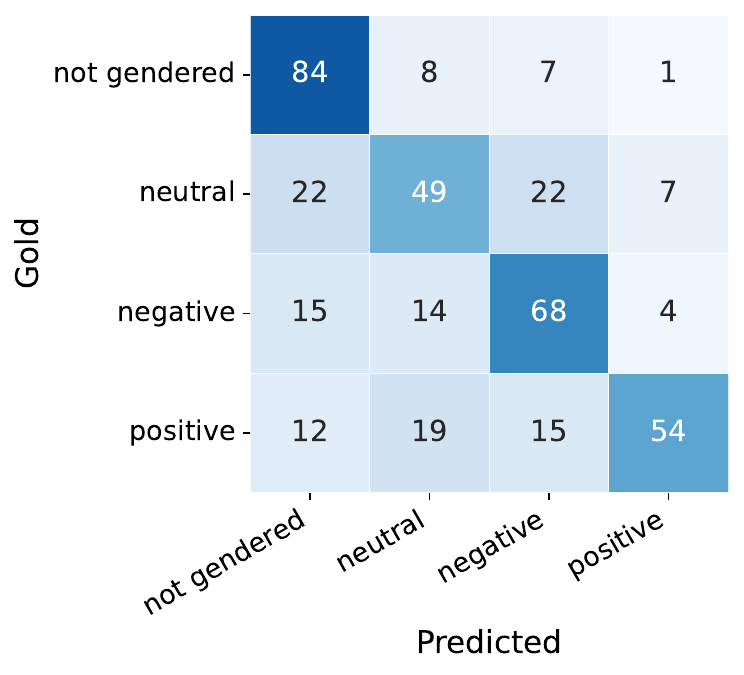}
\caption{End-to-end four-way confusion for SetFit (CV), row-normalized (\%, so each row is recall for a gold class). Neutral gender-related speech is the hardest class, most often misread as negative or not gender-related.}
\label{fig:confusion}
\end{figure}

Figure~\ref{fig:confusion} shows the end-to-end four-way confusion matrix for SetFit (CV), row-normalized so that each row is the recall for a gold class. Neutral gender-related speech is the hardest class, most often misread as gender-negative or not gender-related, as descriptive gender references sit close to both the abusive and the not gender-related classes.

\FloatBarrier
\end{document}